\def\year{2021}\relax
\documentclass[letterpaper]{article}
\usepackage{aaai21} 
\usepackage{times} 
\usepackage{helvet} 
\usepackage{courier} 
\usepackage[hyphens]{url} 
\usepackage{graphicx} 
\usepackage{graphicx} 
\usepackage{natbib}
\usepackage{caption}
\usepackage{amsmath}
\usepackage{amsfonts}
\usepackage{amssymb}
\usepackage{booktabs}
\usepackage{bm}
\usepackage{microtype}

\title{Document-Level Relation Extraction with Adaptive Thresholding and Localized Context Pooling}

\author{
Wenxuan Zhou,\textsuperscript{\rm 1}\thanks{This work was conducted while the first author was doing an internship at JD AI Research.}
Kevin Huang,\textsuperscript{\rm 2}
Tengyu Ma,\textsuperscript{\rm 3}\thanks{TM is also partially supported by the Google Faculty Award, JD.com, Stanford Data Science Initiative, and the Stanford Artificial Intelligence Laboratory.}
Jing Huang \textsuperscript{\rm 2}\\}
\affiliations {
\textsuperscript{\rm 1}Department of Computer Science, University of Southern California, Los Angeles, CA \\
\textsuperscript{\rm 2}JD AI Research, Mountain View, CA\\
\textsuperscript{\rm 3}Department of Computer Science, Stanford University, Stanford, CA\\
zhouwenx@usc.edu,
\{kevin.huang, jing.huang\}@jd.com,
tengyuma@stanford.edu
}
\date{}

\begin{document}
\maketitle
\begin{abstract}
Document-level relation extraction (RE) poses new challenges compared to its sentence-level counterpart.
One document commonly contains multiple entity pairs, and one entity pair occurs multiple times in the document associated with multiple possible relations. 
In this paper, we propose two novel techniques, adaptive thresholding and localized context pooling, to solve the multi-label and multi-entity problems.
The adaptive thresholding replaces the global threshold for multi-label classification in the prior work with a learnable entities-dependent threshold.
The localized context pooling directly transfers attention from pre-trained language models to locate relevant context that is useful to decide the relation. We experiment on three document-level RE benchmark datasets: DocRED, a recently released large-scale RE dataset, and two datasets CDR and GDA in the biomedical domain.
Our ATLOP~(\textbf{A}daptive \textbf{T}hresholding and \textbf{L}ocalized c\textbf{O}ntext \textbf{P}ooling) model achieves an F1 score of $63.4$, and also significantly outperforms existing models on both CDR and GDA.
We have released our code at \url{https://github.com/wzhouad/ATLOP}.
\end{abstract}

\section{Introduction}

Relation extraction (RE) aims to identify the relationship between two entities in a given text and plays an important role in information extraction.
Existing work mainly focuses on sentence-level relation extraction, i.e., predicting the relationship between entities in a single sentence~\cite{Zeng2014RelationCV,Miwa2016EndtoEndRE,Zhang2018GraphCO}.
However, large amounts of relationships, such as relational facts from Wikipedia articles and biomedical literature, are expressed by multiple sentences in real-world applications~\cite{Verga2018SimultaneouslyST,Yao2019DocREDAL}.
This problem, commonly referred to as document-level relation extraction, necessitates models that can capture complex interactions among entities in the whole document.

\begin{figure}[!t]
    \centering
    \includegraphics[width=0.92\linewidth]{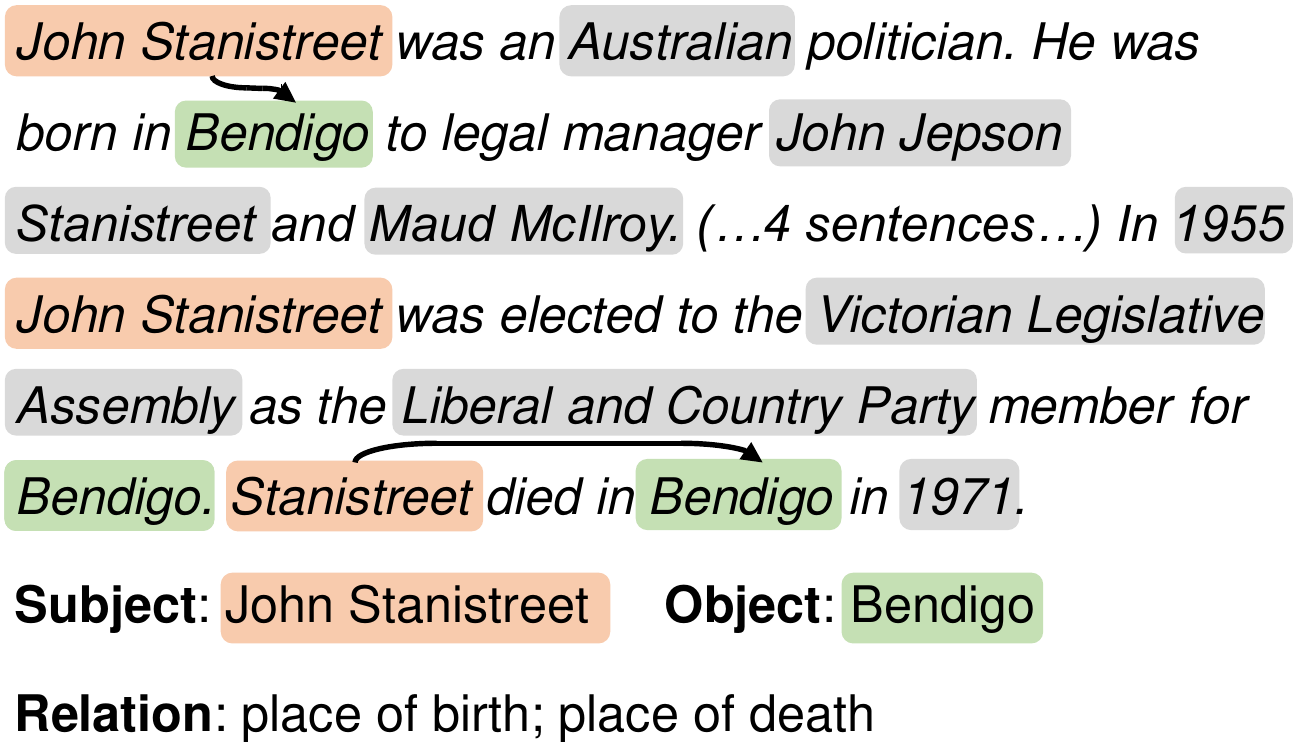}
    \caption{An example of multi-entity and multi-label problems from the DocRED dataset. Subject entity \textit{John Stanistreet} (in orange) and object entity \textit{Bendigo} (in green) express relations \textit{place of birth} and \textit{place of death}. The related entity mentions are connected by lines. Other entities in the document are highlighted in grey.}
    \label{fig:example}
\end{figure}

Compared to sentence-level RE, document-level RE poses unique challenges.
For sentence-level RE datasets such as TACRED~\cite{Zhang2017PositionawareAA} and SemEval 2010 Task 8~\cite{Hendrickx2009SemEval2010T8}, a sentence only contains one entity pair to classify.
On the other hand, for document-level RE, one document contains multiple entity pairs, and we need to classify the relations of them all at once.
It requires the RE model to identify and focus on the part of the document with relevant context for a particular entity pair.
In addition, one entity pair can occur many times in the document associated with distinct relations for document-level RE, in contrast to one relation per entity pair for sentence-level RE.
This multi-entity (multiple entity pairs to classify in a document) and multi-label (multiple relation types for a particular entity pair) properties of document-level relation extraction make it harder than its sentence-level counterpart.
Figure~\ref{fig:example} shows an example from the DocRED dataset~\cite{Yao2019DocREDAL}.
The task is to classify the relation types of pairs of entities (highlighted in color).
For a particular entity pair  (\textit{John Stanistreet}, \textit{Bendigo}), it expresses two relations \textit{place of birth} and \textit{place of death} by the first two sentences and the last sentence.
Other sentences contain irrelevant information to this entity pair.

To tackle the multi-entity problem, most current approaches construct a document graph with dependency structures, heuristics, or structured attention~\cite{Peng2017CrossSentenceNR,Liu2018LearningST,Christopoulou2019ConnectingTD,Nan2020ReasoningWL}, and then perform inference with graph neural models~\cite{Liang2016SemanticOP,Guo2019AttentionGG}.
The constructed graphs bridge entities that spread far apart in the document and thus alleviate the deficiency of RNN-based encoders~\cite{Hochreiter1997LongSM,Chung2014EmpiricalEO} in capturing long-distance information~\cite{Khandelwal2018SharpNF}.
However, as transformer-based models~\cite{Vaswani2017AttentionIA} can implicitly model long-distance dependencies~\cite{Clark2019WhatDB,Tenney2019BERTRT}, it is unclear whether graph structures still help on top of pre-trained language models such as BERT~\cite{Devlin2019BERTPO}.
There have also been approaches to directly apply pre-trained language models without introducing graph structures~\cite{Wang2019FinetuneBF,Tang2020HINHI}.
They simply average the embedding of entity tokens to obtain the entity embeddings and feed them into the classifier to get relation labels.
However, each entity has the same representation in different entity pairs, which can bring noise from irrelevant context.

In this paper, instead of introducing graph structures, we propose a localized context pooling technique.
This technique solves the problem of using the same entity embedding for all entity pairs.
It enhances the entity embedding with additional context that is relevant to the current entity pair.
Instead of training a new context attention layer from scratch, we directly transfer the attention heads from pre-trained language models to get entity-level attention.
Then, for two entities in a pair, we merge their attentions by multiplication to find the context that is important to both of them.

For the multi-label problem, existing approaches reduce it to a binary classification problem.
After training, a global threshold is applied to the class probabilities to get relation labels.
This method involves heuristic threshold tuning and introduces decision errors when the tuned threshold from development data may not be optimal for all instances.

In this paper, we propose the adaptive thresholding technique, which replaces the global threshold with a learnable threshold class.
The threshold class is learned with our adaptive-threshold loss, which is a {\em rank-based} loss that pushes the logits of positive classes above the threshold and pulls the logits of negative classes below in model training.
At the test time, we return classes that have higher logits than the threshold class as the predicted labels or return \textsc{na} if such class does not exist.
This technique eliminates the need for threshold tuning, and also makes the threshold adjustable to different entity pairs, which leads to much better results.

By combining the proposed two techniques, we propose a simple yet effective relation extraction model, named ATLOP (\textbf{A}daptive \textbf{T}hresholding and \textbf{L}ocalized c\textbf{O}ntext \textbf{P}ooling), to fully utilize the power of pre-trained language models~\cite{Devlin2019BERTPO,Liu2019RoBERTaAR}.
This model tackles the multi-label and multi-entity problems in document-level RE.
Experiments on three document-level relation extraction datasets, DocRED~\cite{Yao2019DocREDAL}, CDR~\cite{Li2016BioCreativeVC}, and GDA~\cite{Wu2019RENETAD}, demonstrate that our ATLOP model significantly outperforms the state-of-the-art methods.
The contributions of our work are summarized as follows:
\begin{itemize}
    \item We propose adaptive-thresholding loss, which enables the learning of an adaptive threshold that is dependent on entity pairs and reduces the decision errors caused by using a global threshold.
    \item We propose localized context pooling, which transfers pre-trained attention to grab related context for entity pairs to get better entity representations. 

    \item We conduct experiments on three public document-level relation extraction datasets. Experimental results demonstrate the effectiveness of our ATLOP model that achieves state-of-the-art performance on three benchmark datasets.
\end{itemize}

\section{Problem Formulation}
Given a document $d$ and a set of entities $\{e_i\}_{i=1}^n$, the task of document-level relation extraction is to predict a subset of relations from $\mathcal{R} \cup \{\textsc{na}\}$ between the entity pairs $(e_s, e_o)_{s,\, o=1...n;\, s \neq o}$,
where $\mathcal{R}$ is a pre-defined set of relations of interest, $e_s,e_o$ are identified as subject and object entities, respectively.
An entity $e_i$ can occur multiple times in the document by entity mentions $\{m_j^i\}_{j=1}^{N_{e_i}}$.
A relation exists between entities $(e_s, e_o)$ if it is expressed by any pair of their mentions.
The entity pairs that do not express any relation are labeled \textsc{na}.
At the test time, the model needs to predict the labels of all entity pairs $(e_s, e_o)_{s,\, o=1...n;\, s \neq o}$ in document $d$.

\section{Enhanced BERT Baseline}
In this section, we present our base model for document-level relation extraction.
We build our model based on existing BERT baselines~\cite{Yao2019DocREDAL,Wang2019FinetuneBF} and integrate other techniques to further improve the performance.
\subsection{Encoder}
Given a document $d=[x_t]_{t=1}^l$, we mark the position of entity mentions by inserting a special symbol ``*'' at the start and end of mentions.
It is adapted from the entity marker technique~\cite{Zhang2017PositionawareAA,Shi2019SimpleBM,Soares2019MatchingTB}.
We then feed the document into a pre-trained language model to obtain the contextual embeddings:
\begin{equation}
    \bm{H}=\left[\bm{h}_1, \bm{h}_2, ..., \bm{h}_l\right] = \text{BERT}(\left[x_1, x_2, ..., x_l\right]). \label{eq:encoder}
\end{equation}
Following previous work~\cite{Verga2018SimultaneouslyST,Wang2019ExtractingMI}, the document is encoded once by the encoder, and the classification of all entity pairs is based on the same contextual embedding.
We take the embedding of ``*'' at the start of mentions as the mention embeddings.
For an entity $e_i$ with mentions $\{m_j^i\}_{j=1}^{N_{e_i}}$, we apply logsumexp pooling~\cite{Jia2019DocumentLevelNR}, a smooth version of max pooling, to get the entity embedding $\bm{h}_{e_i}$.
\begin{equation}
    \bm{h}_{e_i} = \log \sum_{j=1}^{N_{e_i}} \exp \left( \bm{h}_{m_j^i} \right). \label{eq:pool}
\end{equation}
This pooling accumulates signals from mentions in the document. It shows better performance compared to mean pooling in experiments.

\subsection{Binary Classifier}
Given the embedding $(\bm{h}_{e_s}, \bm{h}_{e_o})$ of an entity pair $e_s,e_o$ computed by equation~\eqref{eq:pool}, we map the entities to hidden states $\bm{z}$ with a linear layer followed by non-linear activation, then calculate the probability of relation $r$ by bilinear function and sigmoid activation. This process is formulated as:
\begin{align}
    \bm{z}_s &=\tanh \left( \bm{W}_s \bm{h}_{e_s} \right), \label{eq:t1}\\
    \bm{z}_o &=\tanh \left( \bm{W}_o \bm{h}_{e_o} \right), \label{eq:t2}\\
    \mathrm{P}\left(r|e_s, e_o\right) &= \sigma \left(\bm{z}_s^\intercal \bm{W}_r \bm{z}_o + b_r\right), \nonumber
\end{align}
where $\bm{W}_s\in \mathbb{R}^{d \times d}$, $\bm{W}_o\in \mathbb{R}^{d \times d}$, $\bm{W}_r\in \mathbb{R}^{d \times d}, b_r \in \mathbb{R}$ are model parameters.
The representation of one entity is the same among different entity pairs.
To reduce the number of parameters in the bilinear classifier, we use the group bilinear~\cite{Zheng2019LearningDB,Tang2020OTE-ACL}, which splits the embedding dimensions into $k$ equal-sized groups and applies bilinear within the groups:
\begin{align}
    \left[\bm{z}_s^1;...;\bm{z}_s^k \right] &= \bm{z}_s, \nonumber \\
    \left[\bm{z}_o^1;...;\bm{z}_o^k \right] &= \bm{z}_o, \nonumber \\
    \mathrm{P}\left(r|e_s, e_o\right) &= \sigma \left( \sum_{i=1}^k \bm{z}_s^{i\intercal} \bm{W}_r^i \bm{z}_o^i + b_r \right) \label{eq:pred},
\end{align}
where $\bm{W}_r^i\in \mathbb{R}^{d/k \times d/k}$ for $i=1...k$ are model parameters, $\mathrm{P}\left(r|e_s, e_o\right)$ is the probability that relation $r$ is associated with the entity pair $(e_s, e_o)$.
In this way, we can reduce the number of parameters from $d^2$ to $d^2/k$.
We use the binary cross entropy loss for training.
During inference, we tune a global threshold $\theta$ that maximizes evaluation metrics ($F_1$ score for RE) on the development set and return $r$ as an associated relation if $\mathrm{P}\left(r|e_s, e_o\right) > \theta$ or return \textsc{na} if no relation exists.

Our enhanced base model achieves near state-of-the-art performance in our experiments, significantly outperforms existing BERT baselines.

\section{Adaptive Thresholding}
The RE classifier outputs the probability $\mathrm{P}\left(r|e_s, e_o\right)$ within the range $[0,1]$, which needs thresholding to be converted to relation labels.
As the threshold neither has a closed-form solution nor is differentiable, a common practice for deciding threshold is enumerating several values in the range $(0,1)$ and picking the one that maximizes the evaluation metrics ($F_1$ score for RE).
However, the model may have different confidence for different entity pairs or classes in which one global threshold does not suffice.
The number of relations varies (multi-label problem) and the models may not be globally calibrated so that the same probability does not mean the same for all entity pairs.
This problem motivates us to replace the global threshold with a learnable, adaptive one, which can reduce decision errors during inference.

For the convenience of explanation, we split the labels of entity pair $T =(e_s, e_o)$ into two subsets: positive classes $\mathcal{P}_T$ and negative classes $\mathcal{N}_T$, which are defined as follows:
\begin{itemize}
    \item positive classes $\mathcal{P}_T \subseteq \mathcal{R}$ are the relations that exist between the entities in $T$.
    If $T$ does not express any relation, $\mathcal{P}_T$ is empty.
    \item negative classes $\mathcal{N}_T \subseteq \mathcal{R}$ are the relations that do not exist between the entities.
    If $T$ does not express any relation, $\mathcal{N}_T=\mathcal{R}$.
\end{itemize}
If an entity pair is classified correctly, the logits of positive classes should be higher than the threshold while those of negative classes should be lower.
Here we introduce a threshold class $\mathrm{TH}$, which is automatically learned in the same way as other classes (see Eq.\eqref{eq:pred}).
At the test time, we return classes with higher logits than the $\mathrm{TH}$ class as positive classes or return \textsc{na} if such classes do not exist.
This threshold class learns an entities-dependent threshold value.
It is a substitute for the global threshold and thus eliminates the need for tuning threshold on the development set.

\begin{figure}[!t]
    \centering
    \includegraphics[width=0.77\linewidth]{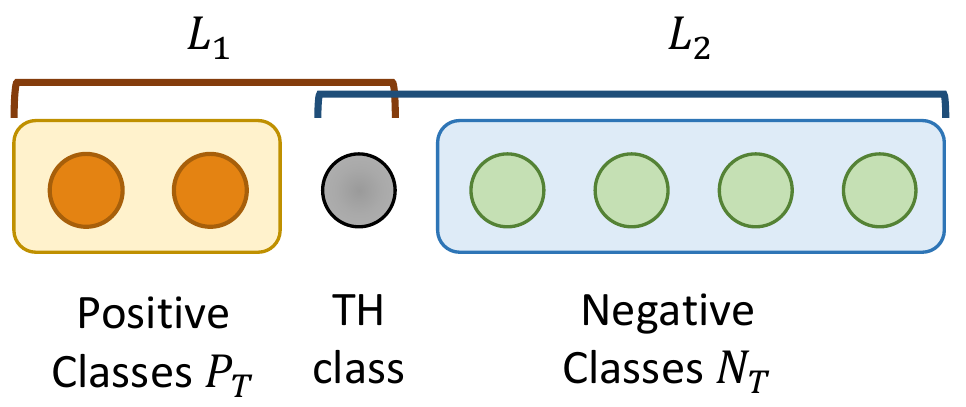}
    \caption{An artificial illustration of our proposed adaptive-thresholding loss. A $\mathrm{TH}$ class is introduced to separate positive classes and negative classes: positive classes would have higher probabilities than $\mathrm{TH}$, and negative classes would have lower probabilities than $\mathrm{TH}$.} 
    \label{fig:atl}
\end{figure}
To learn the new model, we need a special loss function that considers the $\mathrm{TH}$ class.
We design our adaptive-thresholding loss based on the standard categorical cross entropy loss.
The loss function is broken down to two parts as shown below:
\begin{align*}
    \mathcal{L}_1 &= -\sum_{r\in \mathcal{P}_T} \log \left( \frac{\exp\left(\text{logit}_r\right)}{\sum_{r' \in \mathcal{P}_T \cup \{\mathrm{TH}\}} \exp \left(\text{logit}_{r'}\right)} \right), \\
    \mathcal{L}_2 &= - \log \left( \frac{\exp\left(\text{logit}_{\mathrm{TH}}\right)}{\sum_{r' \in \mathcal{N}_T \cup \{\mathrm{TH}\}} \exp \left(\text{logit}_{r'}\right)} \right), \\
    \mathcal{L} &= \mathcal{L}_1 + \mathcal{L}_2.
\end{align*}
The first part $\mathcal{L}_1$ involves positive classes and the $\mathrm{TH}$ class.
Since there may be multiple positive classes, the total loss is calculated as the sum of categorical cross entropy losses on all positive classes~\cite{Menon2019MultilabelRW,Reddi2019StochasticNM}.
$\mathcal{L}_1$ pushes the logits of all positive classes to be higher than the $\mathrm{TH}$ class.
It is not used if there is no positive label.
The second part $\mathcal{L}_2$ involves the negative classes and threshold class.
It is a categorical cross entropy loss with $\mathrm{TH}$ class being the true label.
It pulls the logits of negative classes to be lower than the $\mathrm{TH}$ class.
Two parts are simply summed for the total loss.

The proposed adaptive-thresholding loss is illustrated in Figure~\ref{fig:atl}.
It obtains a large performance gain to the global threshold in our experiments.

\section{Localized Context Pooling}
The logsumexp pooling (see Eq.~\eqref{eq:pool}) accumulates the embedding of all mentions for an entity across the whole document and generates one embedding for this entity.
The entity embedding from this document-level global pooling is then used in the classification of all entity pairs. However, for an entity pair, some context of the entities may not be relevant.
For example, in Figure~\ref{fig:example}, the second mention of \textit{John Stanistreet} and its context are irrelevant to the entity pair (\textit{John Stanistreet}, \textit{Bendigo}).
Therefore, it is better to have a localized representation that only attends to the relevant context in the document that is useful to decide the relation for this entity pair.

\begin{figure}[!t]
    \centering
    \includegraphics[width=0.9\linewidth]{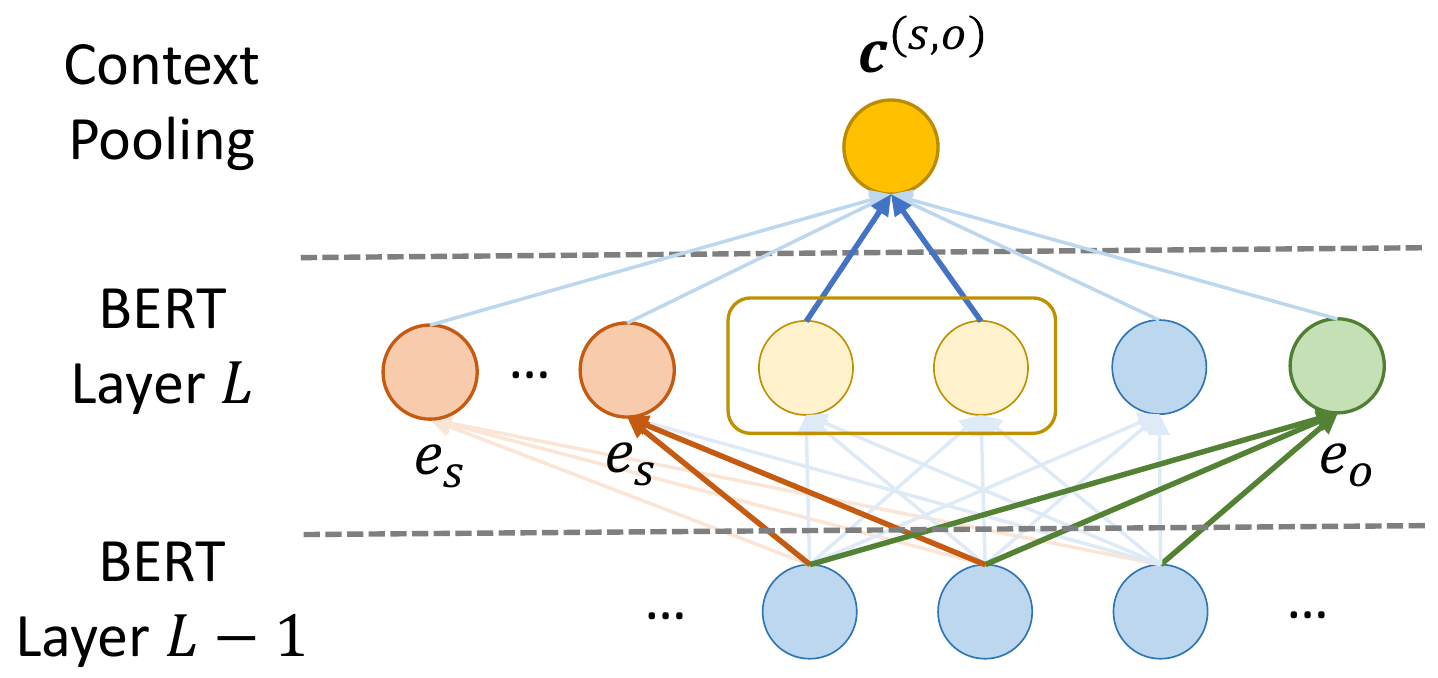}
    \caption{\textbf{Illustration of localized context pooling.} Tokens are weighted averaged to form the localized context $\bm{c}^{(s, o)}$ of the entity pair $(e_s, e_o)$.
    The weights of tokens are derived by multiplying the attention weights of the subject entity $e_s$ and the object entity $e_o$ from the last transformer layer so that only the tokens that are important to both entities (highlighted in light yellow) receive higher weights.
    }
    \label{fig:lcp}
\end{figure}

Therefore we propose the localized context pooling, where we enhance the embedding of an entity pair with an additional local context embedding that is related to both entities.
In this work, since we use pre-trained transformer-based models as the encoder, which has already learned token-level dependencies by multi-head self-attention~\cite{Vaswani2017AttentionIA}, we consider directly using their attention heads for localized context pooling.
This method transfers the well-learned dependencies from the pre-trained language model without learning new attention layers from scratch.

Specifically, given a pre-trained multi-head attention matrix $\bm{A}\in \mathbb{R}^{H \times l \times l}$, where $\bm{A}_{ijk}$ represents attention from token $j$ to token $k$ in the $i^{th}$ attention head, we first take the attention from the ``*'' symbol as the mention-level attention, then average the attention over mentions of the same entity to obtain entity-level attention $\bm{A}^E_i\in \mathbb{R}^{H\times l}$, which denotes attention from the $i^{th}$ entity to all tokens.
Then given an entity pair $(e_s, e_o)$, we locate the local context that is important to both $e_s$ and $e_o$ by multiplying their entity-level attention, and obtain the localized context embedding $\bm{c}^{(s,o)}$ by:
\begin{align*}
    \bm{A}^{(s,o)} &= \bm{A}^E_s \cdot \bm{A}^E_o, \\
    \bm{q}^{(s,o)} &= \sum_{i=1}^H \bm{A}^{(s,o)}_i, \\
    \bm{a}^{(s,o)} &= \bm{q}^{(s,o)} / {\bm{1}^\intercal \bm{q}^{(s,o)}}, \\
    \bm{c}^{(s,o)} &= \bm{H}^\intercal \bm{a}^{(s,o)},
\end{align*}
where $\bm{H}$ is the contextual embedding in Eq.~\eqref{eq:encoder}.
The localized context embedding is then fused into the globally pooled entity embedding to obtain entity representations that are different for different entity pairs, by modifying the original linear layer in Eq.~\eqref{eq:t1} and Eq.~\eqref{eq:t2} as follows:
\begin{align}
    \bm{z}_s^{(s,o)} &=\tanh \left( \bm{W}_s \bm{h}_{e_s} + \bm{W}_{c_1} \bm{c}^{(s,o)} \right), \label{eq:t5} \\
    \bm{z}_o^{(s,o)} &=\tanh \left( \bm{W}_o \bm{h}_{e_o} + \bm{W}_{c_2} \bm{c}^{(s,o)} \right), \label{eq:t6} 
\end{align}
where $\bm{W}_{c_1}, \bm{W}_{c_2} \in \mathbb{R}^{d \times d}$ are model parameters.
The proposed localized context pooling is illustrated in Figure~\ref{fig:lcp}.
In experiments, we use the attention matrix from the last transformer layer.

\section{Experiments}

\begin{table}[!t]
\centering
\scalebox{0.88}{
    \begin{tabular}{p{4cm}ccc}
         \toprule
         \textbf{Statistics} & DocRED & CDR & GDA \\
         \midrule
         \# Train& 3053& 500&  23353 \\
         \# Dev& 1000& 500& 5839 \\
         \# Test& 1000& 500& 1000 \\
         \# Relations& 97& 2& 2 \\
         Avg.\# entities per Doc.& 19.5& 7.6& 5.4\\
         \bottomrule
    \end{tabular}}
    \caption{Statistics of the datasets in experiments.}
    \label{tab::statistics}
\end{table}

\begin{table}[!t]
\centering
\scalebox{0.88}{
    \begin{tabular}{p{2cm}cccc}
         \toprule
         \textbf{Hyperparam}& \multicolumn{2}{c}{\textbf{DocRED}} & \multicolumn{1}{c}{\textbf{CDR}}& \multicolumn{1}{c}{\textbf{GDA}} \\
         & BERT& RoBERTa & SciBERT & SciBERT \\
         \midrule
         Batch size& 4& 4 & 4& 16 \\
         \# Epoch& 30&30&30&10 \\
         lr for encoder&5e-5&3e-5&2e-5&2e-5  \\
         lr for classifier& 1e-4& 1e-4& 1e-4& 1e-4 \\
         \bottomrule
    \end{tabular}}
    \caption{Hyper-parameters in training.}
    \label{tab::hyper}
\end{table}

\begin{table*}[!t]
\centering
\scalebox{0.88}{
    \begin{tabular}{p{6.5cm}cccc}
         \toprule
         \textbf{Model} & \multicolumn{2}{c}{\textbf{Dev}} & \multicolumn{2}{c}{\textbf{Test}} \\
          & Ign $F_1$ & $F_1$ & Ign $F_1$ & $F_1$ \\
         \midrule
         \emph{Sequence-based Models} \\
         CNN~\cite{Yao2019DocREDAL}& 41.58& 43.45& 40.33& 42.26 \\
         BiLSTM~\cite{Yao2019DocREDAL}& 48.87& 50.94& 48.78& 51.06 \\
         \midrule
         \emph{Graph-based Models} \\
         BiLSTM-AGGCN~\cite{Guo2019AttentionGG}& 46.29& 52.47& 48.89& 51.45 \\
         BiLSTM-LSR~\cite{Nan2020ReasoningWL}& 48.82& 55.17& 52.15& 54.18 \\
         BERT-LSR$_{\text{BASE}}$~\cite{Nan2020ReasoningWL}& 52.43& 59.00& 56.97& 59.05\\
         \midrule
         \emph{Transformer-based Models} \\
         BERT$_{\text{BASE}}$~\cite{Wang2019FinetuneBF}& -& 54.16& -& 53.20 \\
         BERT-TS$_{\text{BASE}}$~\cite{Wang2019FinetuneBF}& -& 54.42&-& 53.92 \\
         HIN-BERT$_{\text{BASE}}$~\cite{Tang2020HINHI}& 54.29& 56.31& 53.70& 55.60 \\
         CorefBERT$_{\text{BASE}}$~\cite{Ye2020CoreferentialRL}& 55.32& 57.51& 54.54& 56.96 \\
         CorefRoBERTa$_{\text{LARGE}}$~\cite{Ye2020CoreferentialRL}& 57.35& 59.43& 57.90& 60.25 \\
         \midrule
         \emph{Our Methods} \\
         BERT$_{\text{BASE}}$ (our implementation)& 54.27 $\pm$ 0.28& 56.39 $\pm$ 0.18& -&- \\
         BERT-E$_{\text{BASE}}$& 56.51 $\pm$ 0.16& 58.52 $\pm$ 0.19& -&-\\
         BERT-ATLOP$_{\text{BASE}}$& 59.22 $\pm$ 0.15& 61.09 $\pm$ 0.16& 59.31& 61.30 \\
         RoBERTa-ATLOP$_{\text{LARGE}}$& \textbf{61.32 $\pm$ 0.14}& \textbf{63.18 $\pm$ 0.19}& \textbf{61.39}& \textbf{63.40} \\
         \bottomrule
    \end{tabular}}
    \caption{Main results (\%) on the development and test set of DocRED. We report the mean and standard deviation of $F_1$ on the development set by conducting 5 runs of training using different random seeds. We report the official test score of the best checkpoint on the development set.}
    \label{tab::main_results}
\end{table*}

\begin{table}[!t]
\centering
\scalebox{0.88}{
    \begin{tabular}{p{5cm}cc}
         \toprule
         \textbf{Model} & CDR& GDA \\
         \midrule
         BRAN~\cite{Verga2018SimultaneouslyST}& 62.1& - \\
         CNN~\cite{Nguyen2018ConvolutionalNN}& 62.3& - \\
         EoG~\cite{Christopoulou2019ConnectingTD}& 63.6& 81.5 \\
         LSR~\cite{Nan2020ReasoningWL}& 64.8& 82.2\\
         \midrule
         SciBERT (our implementation)& 65.1 $\pm$ 0.6& 82.5 $\pm$ 0.3 \\
         SciBERT-E& 65.9 $\pm$ 0.5& 83.3 $\pm$ 0.3 \\
         SciBERT-ATLOP& \textbf{69.4 $\pm$ 1.1}& \textbf{83.9 $\pm$ 0.2} \\
         \bottomrule
    \end{tabular}}
    \caption{Test $F_1$ score (\%) on CDR and GDA dataset. Our ATLOP model with the SciBERT encoder outperforms the current SOTA results.
    }
    \label{tab::bio_result}
\end{table}

\subsection{Datasets}
We evaluate our ATLOP model on three public document-level relation extraction datasets.
The dataset statistics are shown in Table~\ref{tab::statistics}.
\begin{itemize}
    \item \textbf{DocRED}~\cite{Yao2019DocREDAL} is a large-scale crowdsourced dataset for document-level RE.
    It is constructed from Wikipedia articles.
    DocRED consists of 3053 documents for training.
    For entity pairs that express relation(s), about 7\% of them have more than one relation label.
    \item \textbf{CDR}~\cite{Li2016BioCreativeVC} is a human-annotated dataset in the biomedical domain.
    It consists of 500 documents for training.
    The task is to predict the binary interactions between Chemical and Disease concepts.
    \item \textbf{GDA}~\cite{Wu2019RENETAD} is a large-scale dataset in the biomedical domain.
    It consists of 29192 articles for training.
    The task is to predict the binary interactions between Gene and Disease concepts.
    We follow \citet{Christopoulou2019ConnectingTD} to split the training set into an 80/20 split as training and development sets.
\end{itemize}
\subsection{Experiment Settings}
Our model is implemented based on Huggingface's Transformers~\cite{wolf2019huggingface}.
We use cased BERT-base~\cite{Devlin2019BERTPO} or RoBERTa-large~\cite{Liu2019RoBERTaAR} as the encoder on DocRED, and cased SciBERT~\cite{Beltagy2019SciBERTAP} on CDR and GDA.
We use mixed-precision training~\cite{Micikevicius2018MixedPT} based on the Apex library\footnote{\url{https://github.com/NVIDIA/apex}}.
Our model is optimized with AdamW~\cite{Loshchilov2019DecoupledWD} using learning rates $\in \{2\mathrm{e}{-5}, 3\mathrm{e}{-5}, 5\mathrm{e}{-5}, 1\mathrm{e}{-4}\}$, with a linear warmup~\cite{Goyal2017AccurateLM} for the first 6\% steps followed by a linear decay to 0.
We apply dropout~\cite{Srivastava2014DropoutAS} between layers with rate 0.1, and clip the gradients of model parameters to a max norm of 1.0.
We perform early stopping based on the $F_1$ score on the development set.
All hyper-parameters are tuned on the development set.
We list some of the hyper-parameters in Table~\ref{tab::hyper}.

For models that use a global threshold, we search threshold values from $\{0.1, 0.2, ..., 0.9\}$ and pick the one that maximizes dev $F_1$.
All models are trained with 1 Tesla V100 GPU.
For the DocRED dataset, the training takes about 1 hour 45 minutes with BERT-base encoder and 3 hours 30 minutes with RoBERTa-large encoder.
For CDR and GDA datasets, the training takes 20 minutes and 3 hours 30 minutes with SciBERT encoder, respectively.
\subsection{Main Results}
We compare ATLOP with sequence-based models, graph-based models, and transformer-based models on the DocRED dataset.
The experiment results are shown in Table~\ref{tab::main_results}.
Following \citet{Yao2019DocREDAL}, we use $F_1$ and Ign $F_1$ in evaluation.
The Ign $F_1$ denotes the $F_1$ score excluding the relational facts that are shared by the training and dev/test sets.

\smallskip
\noindent
\textbf{Sequence-based Models.}
These models use neural architectures such as CNN~\cite{Goodfellow2015DeepL} and bidirectional LSTM~\cite{Schuster1997BidirectionalRN} to encode the entire document, then obtain entity embeddings and predict relations for each entity pair with bilinear function.

\smallskip
\noindent
\textbf{Graph-based Models.} 
These models construct document graphs by learning latent graph structures of the document and perform inference with graph convolutional network~\cite{Kipf2017SemiSupervisedCW}.
We include two state-of-the-art graph-based models, AGGCN~\cite{Guo2019AttentionGG} and LSR~\cite{Nan2020ReasoningWL}, for comparison.
The result of AGGCN is from the re-implementation by \citet{Nan2020ReasoningWL}.

\smallskip
\noindent
\textbf{Transformer-based Models.}
These models directly adapt pre-trained language models to document-level RE without using graph structures.
They can be further divided into pipeline models (BERT-TS~\cite{Wang2019FinetuneBF}), hierarchical models (HIN-BERT~\cite{Tang2020HINHI}), and pre-training methods (CorefBERT and CorefRoBERTa~\cite{Ye2020CoreferentialRL}).
We also include the BERT baseline~\cite{Wang2019FinetuneBF} and our re-implemented BERT baseline in comparison.

We find that our re-implemented BERT baseline gets significantly better results than \citet{Wang2019FinetuneBF}, and outperforms the state-of-the-art RNN-based model BiLSTM-LSR by $1.2\%$.
It demonstrates that pre-trained language models can capture long-distance dependencies among entities without explicitly using graph structures.
After integrating other techniques, our enhanced baseline BERT-E$_{\text{BASE}}$ achieves an F1 score of $58.52\%$, which is close to the current state-of-the-art model BERT-LSR$_{\text{BASE}}$.
Our BERT-ATLOP$_{\text{BASE}}$ model further improves the performance of BERT-E$_{\text{BASE}}$ by $2.6\%$, demonstrating the efficacy of the proposed two novel techniques.
Using RoBERTa-large as the encoder, our ALTOP model achieves an F1 score of $63.40\%$, which is a new state-of-the-art result on DocRED.

\subsection{Results on Biomedical Datasets}
Experiment results on two biomedical datasets are shown in Table~\ref{tab::bio_result}.
~\citet{Verga2018SimultaneouslyST} and ~\citet{Nguyen2018ConvolutionalNN} are both sequence-based models that use self-attention network and CNN as the encoders, respectively.
~\citet{Christopoulou2019ConnectingTD} and ~\citet{Nan2020ReasoningWL} use graph-based models that construct document graphs by heuristics or structured attention, and perform inference with graph neural network.
To our best knowledge, transformer-based pre-trained language models have not been applied to document-level RE datasets in the biomedical domain.
In experiments, we replace the encoder with SciBERT, which is pre-trained on multi-domain corpora of scientific publications.
The SciBERT baseline already outperforms all existing methods.
Our SciBERT-ATLOP model further improves the $F_1$ score by $4.3\%$ and $1.4\%$ on CDR and GDA, respectively, yielding new state-of-the-art results on these two datasets.

\begin{table}[!t]
\centering
\scalebox{0.9}{
    \begin{tabular}{p{4.5cm}cc}
         \toprule
         \textbf{Model} & Ign $F_1$ & $F_1$ \\
         \midrule
         BERT-ATLOP$_{\text{BASE}}$& 59.22& 61.09 \\
         $-$ Adaptive Thresholding& 58.32& 60.20 \\
         $-$ Localized Context Pooling& 58.19& 60.12\\
         $-$ Adaptive-Thresholding Loss& 39.52& 41.74\\
         \midrule
         BERT-E$_{\text{BASE}}$& 56.51& 58.52 \\
         $-$ Entity Marker& 56.22& 58.28 \\
         $-$ Group Bilinear& 55.51& 57.54 \\
         $-$ Logsumexp Pooling& 55.35& 57.40 \\
         \bottomrule
    \end{tabular}}
    \caption{Ablation study of ATLOP on DocRED. We turn off different components of the model one at a time. 
    These ablation results show that both adaptive thresholding and localized context pooling are effective. Logsumexp pooling and group bilinear both bring noticeable gain to the baseline.
    }
    \label{tab::ablation}
\end{table}

\subsection{Ablation Study}
To show the efficacy of our proposed techniques, we conduct two sets of ablation studies on ATLOP and enhanced baseline, by turning off one component at a time.
We observe that all components contribute to model performance.
The adaptive thresholding and localized context pooling are equally important to model performance, leading to a drop of $0.89\%$ and $0.97\%$ in dev $F_1$ score respectively when removed from ATLOP.
Note that the adaptive thresholding only works when the model is optimized with the adaptive-thresholding loss.
Applying adaptive thresholding to models trained with binary cross entropy results in dev $F_1$ of $41.74\%$. 

For our enhanced baseline model BERT-E$_{\text{BASE}}$, both group bilinear and logsumexp pooling lead to about $1\%$ increase in dev $F_1$.
We find the improvement from entity markers is minor ($0.24\%$ in dev $F_1$) but still use the technique in the model as it makes the derivation of mention embedding and mention-level attention easier.

\begin{table}[!t]
\centering
\scalebox{0.87}{
    \begin{tabular}{p{4cm}cc}
         \toprule
         \textbf{Strategy} & Dev $F_1$ & Test $F_1$ \\
         \midrule
         Global Thresholding& 60.14& 60.62 \\
         Per-class Thresholding& \textbf{61.73}& 60.35 \\
         Adaptive Thresholding& 61.27& \textbf{61.30} \\
         \bottomrule
    \end{tabular}}
    \caption{Result of different thresholding strategies on DocRED. Our adaptive thresholding consistently outperforms other strategies on the test set.}
    \label{tab::thresholding}
\end{table}

\subsection{Analysis of Thresholding}
Global thresholding does not consider the variations of model confidence in different classes or instances, and thus yields suboptimal performance.
One interesting question is whether we can improve global thresholding by tuning different thresholds for different classes.
To answer this question, We try to tune different thresholds on different classes to maximize the dev $F_1$ score on DocRED using the cyclic optimization algorithm~\cite{Fan2007ASO}.
Results are shown in Table~\ref{tab::thresholding}.
We find that using per-class thresholding significantly improves the dev $F_1$ score to $61.73\%$, which is even higher than the result of adaptive thresholding.
However, this gain does not transfer to the test set.
The result of per-class thresholding is even worse than global thresholding.
It indicates that tuning per-class thresholding after training can lead to severe over-fitting to the development set.
While our adaptive thresholding technique learns the threshold in training, which can generalize to the test set.

\begin{figure}[!t]
    \centering
    \includegraphics[width=0.9\linewidth]{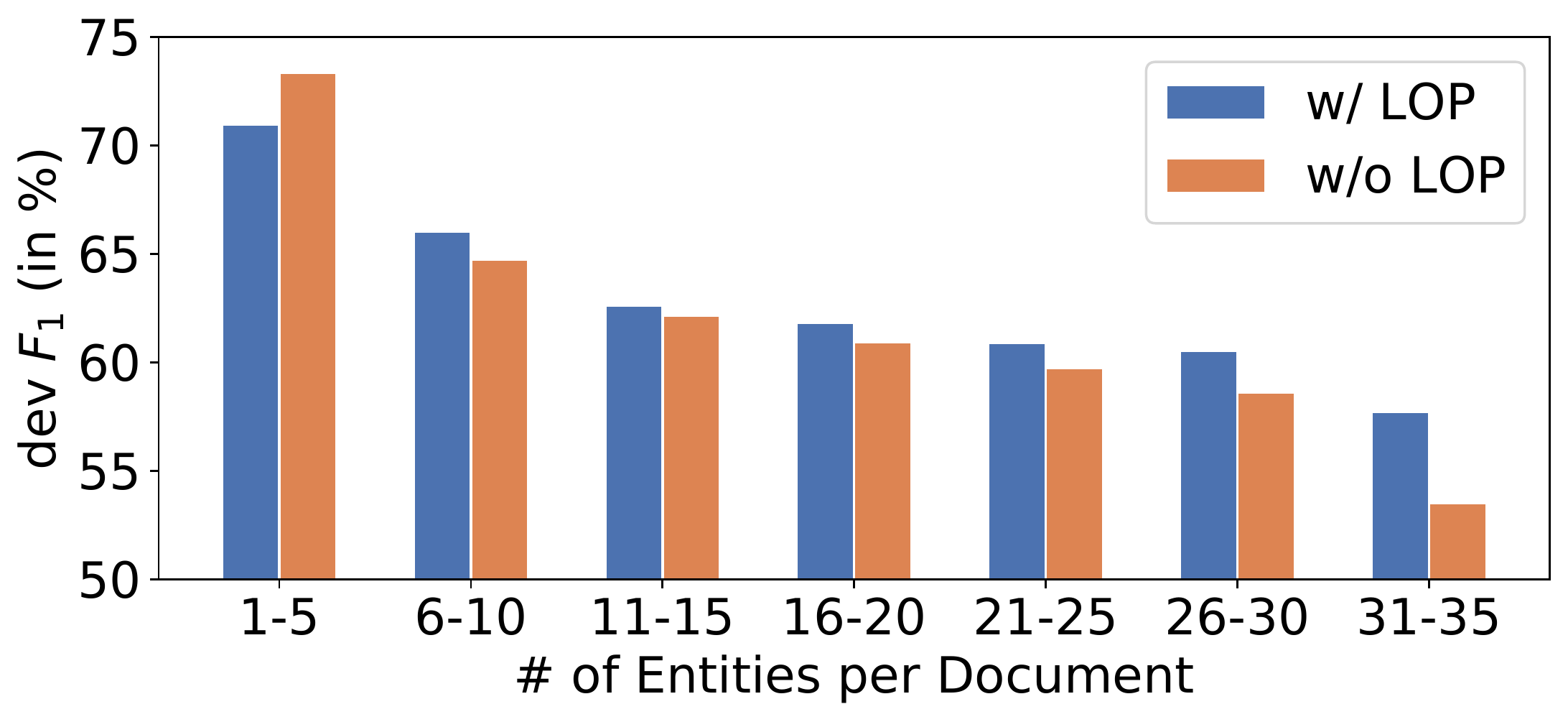}
    \caption{Dev $F_1$ score of documents with the different number of entities on DocRED. Our localized context pooling achieves better results when the number of entities is larger than 5. The improvement becomes more significant when the number of entities increases.
    }
    \label{fig:lop_f1}
\end{figure}

\subsection{Analysis of Context Pooling}
To show that our localized context pooling~(LOP) technique mitigates the multi-entity issue, we divide the documents in the development set of DocRED into different groups by the number of entities, and evaluate models trained with or without localized context pooling on each group.
Experiment results are shown in Figure~\ref{fig:lop_f1}.
We observe that for both models, their performance gets worse when the document contains more entities.
The model w/ LOP consistently outperforms the model w/o LOP except when the document contains very few entities (1 to 5), and the improvement gets larger when the number of entities increases.
However, the number of documents that only contain 1 to 5 entities is very small (4 in the dev set), and the documents in DocRED contain 19 entities on average. 
Therefore our localized context pooling still improves the overall $F_1$ score significantly.
This indicates that the localized context pooling technique can capture related context for entity pairs and thus alleviates the multi-entity problem.

\begin{figure}[!t]
    \centering
    \includegraphics[width=0.95\linewidth]{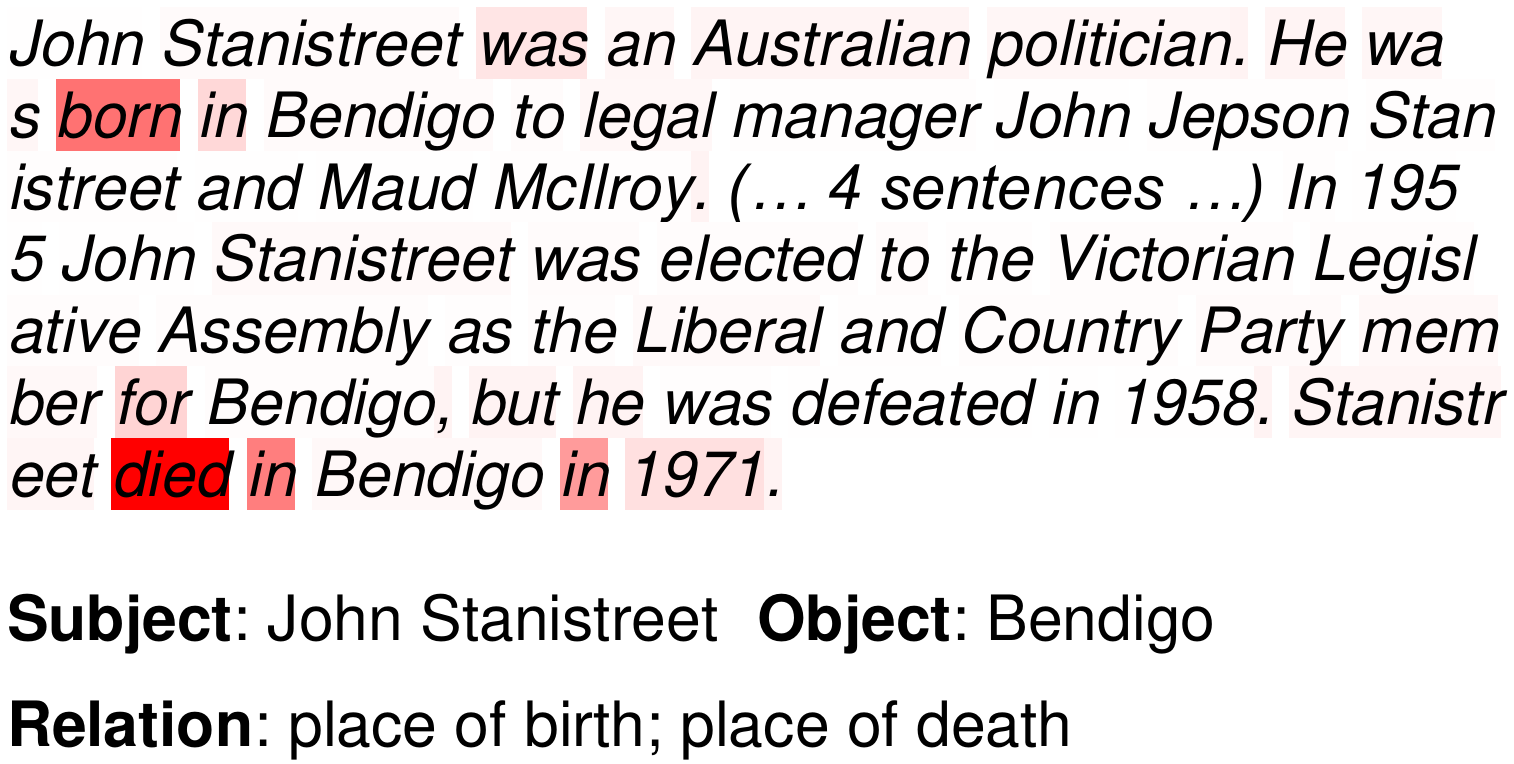}
    \caption{Context weights of an example from DocRED. We visualize the weight of context tokens $\bm{a}^{(s, o)}$ in localized context pooling. The model attends to the most relevant context \textit{born} and \textit{died} for entity pair (\textit{John Stanistreet}, \textit{Bendigo}).
    }
    \label{fig:attention}
\end{figure}

We also visualize the context weights of the example in Figure~\ref{fig:example}.
As shown in Figure~\ref{fig:attention}, our localized context pooling gives high weights to \textit{born} and \textit{died}, which are most relevant to both entities (\textit{John Stanistreet}, \textit{Bendigo}).
These two tokens are also evidence for the two ground truth relationships \textit{place of birth} and \textit{place of death}, respectively.
Tokens like \textit{elected} and \textit{politician} get much smaller weights because they are only related to the subject entity \textit{John Stanistreet}.
The visualization demonstrates that the localized context can locate the context that is related to both entities.

\section{Related Work}
Early research efforts on relation extraction concentrate on predicting the relationship between two entities within a sentence.
Various approaches including sequence-based methods~\cite{Zeng2014RelationCV,Wang2016RelationCV,Zhang2017PositionawareAA}, graph-based methods~\cite{Miwa2016EndtoEndRE,Zhang2018GraphCO,Guo2019AttentionGG,Wu2019SimplifyingGC}, transformer-based methods~\cite{Alt2019ImprovingRE,Shi2019SimpleBM}, and pre-training methods~\cite{Zhang2019ERNIEEL,Soares2019MatchingTB} have been shown effective in tackling this problem.

However, as large amounts of relationships are expressed by multiple sentences~\cite{Verga2018SimultaneouslyST,Yao2019DocREDAL}, recent work starts to explore document-level relation extraction.
Most approaches on document-level RE are based on document graphs, which were introduced by \citet{Quirk2017DistantSF}.
Specifically, they use words as nodes and inner and inter-sentential dependencies (dependency structures, coreferences, etc.) as edges.
This document graph provides a unified way of extracting the features for entity pairs.
Later work extends the idea by improving neural architectures~\cite{Peng2017CrossSentenceNR,Verga2018SimultaneouslyST,Song2018NaryRE,Jia2019DocumentLevelNR,Gupta2019NeuralRE} or adding more types of edges~\cite{Christopoulou2019ConnectingTD,Nan2020ReasoningWL}.
In particular, \citet{Christopoulou2019ConnectingTD} constructs nodes of different granularities (sentence, mention, entity), connects them with heuristically generated edges, and infers the relations with an edge-oriented model~\cite{Christopoulou2018AWM}.
\citet{Nan2020ReasoningWL} treats the document graph as a latent variable and induces it by structured attention~\cite{Liu2018LearningST}.
This work also proposes a refinement mechanism to enable multi-hop information aggregation from the whole document.
Their LSR model achieved state-of-the-art performance on document-level RE.

There have also been models that directly apply pre-trained language models without introducing document graphs, since edges such as dependency structures and coreferences can be automatically learned by pre-trained language models~\cite{Clark2019WhatDB,Tenney2019BERTRT,Vig2019AnalyzingTS,Hewitt2019ASP}.
In particular, \citet{Wang2019FinetuneBF} proposes a pipeline model that first predicts whether a relationship exists in an entity pair and then predicts the specific relation types.
\citet{Tang2020HINHI} proposes a hierarchical model that aggregates entity information from the entity level, sentence level, and document level.
\citet{Ye2020CoreferentialRL} introduces a copy-based training objective to pre-training, which enhances the model's ability in capturing coreferential information and brings noticeable gain on various NLP tasks that require coreferential reasoning.

However, none of the models focus on the multi-entity and multi-label problems, which are among the key differences of document-level RE to its sentence-level RE counterpart.
Our ATLOP model deals with the two problems by two novel techniques: adaptive thresholding and localized context pooling, and significantly outperforms existing models.

\section{Conclusion}
In this work, we propose the ATLOP model for document-level relation extraction, which features two novel techniques: adaptive thresholding and localized context pooling.
The adaptive thresholding technique replaces the global threshold in multi-label classification with a learnable threshold class that can decide the best threshold for each entity pair.
The localized context pooling utilizes pre-trained attention heads to locate relevant context for entity pairs and thus helps in alleviating the multi-entity problem.
Experiments on three public document-level relation extraction datasets demonstrate that our ATLOP model significantly outperforms existing models and yields the new state-of-the-art results on all datasets.

\bibliography{aaai21}
\end{document}